\documentclass[conference,a4paper]{ieeetran}

\usepackage{amsfonts,amsmath,amssymb}
\usepackage{graphicx}
\usepackage{algorithmic, algorithm}
\usepackage{color,colortbl,flushend}
\usepackage[bookmarks=false]{hyperref}

\newcommand{\CL}{\cellcolor[gray]{0.8}}

\usepackage{fancyheadings}
\title{Combining Real-Valued and Binary Gabor-Radon Features for Classification and Search in Medical Imaging Archives}

\author{\authorblockN{Hamed Erfankhah$^\dagger$, Mehran Yazdi$^\dagger$, H.R.Tizhoosh$^\ddagger$}
\authorblockA{$^\dagger$Electrical and Computer Engineering, Shiraz University, Shiraz, Iran\\ erfankhah.hamed@shirazu.ac.ir, yazdi@shirazu.ac.ir  }
\authorblockA{$^\ddagger$KIMIA Lab, University of Waterloo, Ontario,
Canada, \url{http://kimia.uwaterloo.ca/}} }

\begin{document}
\maketitle

\begin{abstract}
Content-based image retrieval (CBIR) of medical images in large datasets to identify similar images when a query image is given can be very useful in improving the diagnostic decision of the clinical experts and as well in educational scenarios. In this paper, we used two stage classification and retrieval approach to retrieve similar images. First, the Gabor filters are applied to Radon-transformed images to extract features and to train a multi-class SVM. Then based on the classification results and using an extracted Gabor barcode, similar images are retrieved. The proposed method was tested on IRMA dataset which contains more than 14,000 images. Experimental results show the efficiency of our approach in retrieving similar images compared to other Gabor-Radon-oriented methods.
\end{abstract}

\section{Introduction}
With the rapid development of digital imaging as well as the widespread usage of picture archiving systems, the problem of efficiently retrieving relevant information from big data has become one of the most interesting research topics in recent years. Digital images are widely used in various fields especially in medicine. Images provide a fast and non-invasive way for diagnosis, treatment planning, and monitoring of many diseases. Digital imagery can also be useful in the education domain to improve the theoretical and experimental knowledge of students and residents \cite{1}. 

Searching for images via textual information (the conventional image search as we know from the World Wide Web) cannot be employed for medical imaging. Manually annotating medical images by experts is extremely time-consuming. As well, annotations cannot describe the full image content (e.g., the irregular shape of a tumour) by some textual descriptions to discriminate the similarities or differences between diverse categories. So retrieving images by contents, or content-based image retrieval (CBIR), has been a growing field for more than two decades. The aim of CBIR is to index and to retrieve the images via their visual contents such as color, texture, shape or their combination \cite{2,3}. In a typical CBIR system, the user submits a query image to the system. After that, a feature extraction method is applied to the query image to create a feature vector. This feature vector can be compared with the feature vectors stored in the database through similarity calculation. The most similar images with respect to the query image are then retrieved by picking the stored case with the shortest distance. Feature extraction is the main step of CBIR systems; the content of the image is quantified in a vector of numbers, a task that arguably is paramount for the retrieval of relevant information.  

In this paper, a two stage classification and retrieval approach is used. We applied Radon and Gabor filters to extract image features and to train a multi-class SVM classifier. In the retrieval stage, the Gabor-Radon barcodes are used to retrieve the similar images from within the class determined by the SVM. 

The rest of the paper is organized as follows: Section II gives a brief overview of relevant literature. Section III provides a short introduction of Radon transform, Radon barcodes, Gabor filter bank and support vector machine. In section IV, we describe our method. In section V, we report the experiments and results. Section VI concludes the paper. 

\section{Image Retrieval}
From the beginning years of research on medical image retrieval, various searching methods have been proposed. The retrieval of medical images in Picture Archival and Communication System (PACS), which provides convenient access to images of different modalities in a DICOM (Digital Imaging and Communications in Medicine) format, is of great importance.

Tommasi et al. used support vector machines (SVMs) to tag medical images by combining global and local features based on a multi-cue approach and achieved good results on the ImageCLEF 2009 medical image annotation task \cite{6,7}. In another work, a hierarchical multi-label classification (HMC) system has been used for medical image tagging \cite{8}. Gabor filter banks have been used typically to extract textural features for image medical retrieval \cite{4}. Recently, a noticeable trend in CBIR research is to extract binary features, sometimes called \emph{barcodes}, because binary features are fast in processing and also require less storage space. Leutengger et al. used Binary Robust Invariant Scalable keypoints (BRISK) for image keypoint detection, description and matching \cite{9}. Radon barcodes based on Radon transform for encoding the local image information was introduced recently \cite{10}. Using Gabor filters to extract binary codes from Radon transformed image was proposed by Nouresanesh et al. \cite{5,5a}. Camlica et al. used an SVM classifier trained with LBP features derived from saliency image regions \cite{11}. Extracting center-symmetric local binary pattern (LBP) of the image and then computing the gray level co-occurrence matrix to retrieval propose has been used as well \cite{12}. Zhu et al. combined Radon projections and SVM to retrieve medical images \cite{13}.

Most recently, deep architectures have also been used for medical image retrieval. Deep denoising autoencoder (DDA), for instance, was used to hash the X-ray images into binary codes \cite{14}. Deep convolutional based image retrieval neural networks have also been investigated \cite{15,16}. The drawback of such solutions is that they require a large, labeled and balanced dataset, and a lot of computational resources for training.

\section{Background Review }
\subsection{Radon Transform}
\begin{figure}[!t]
\begin{center}
\includegraphics[width=0.9\columnwidth]{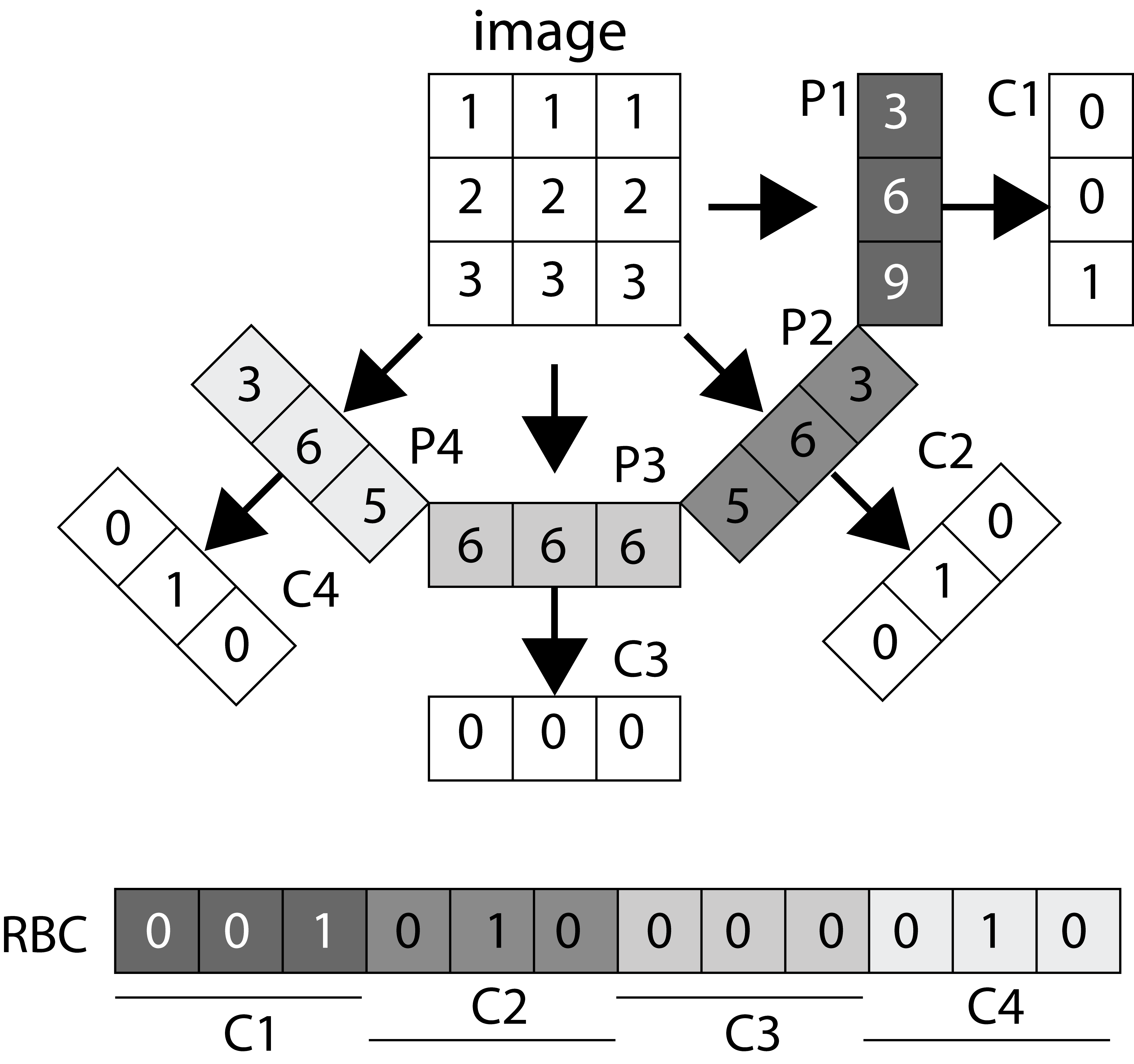}
\caption{Generating Radon barcodes \cite{10}.}
\label{fig:RBC}
\end{center}
\end{figure}

 The Radon transform is an integral transform, which takes the function  and computes the projection of it along various directions. The Radon transform was introduced by J.Radon in 1917 \cite{17}.  Due to the inherent properties of the Radon transform such as rotation invariance and robustness to zero mean white noise, it has been applied in many applications like computed axial tomography, categorizing visual objects \cite{18}, and object detection  \cite{19}. The Radon transform of the image $I$ as a 2D function, which is a new image $R(\rho,\theta)$, is its line integral along a line inclined at an angle $\theta$ and at a distance $\rho$ from the origin as follows: 
\begin{equation}
R(\rho,\theta)=\int\limits_{-\infty}^{+\infty}\int\limits_{-\infty}^{+\infty} f(x,y)\delta(\rho-x\cos(\theta)-y\sin(\theta)) dxdy,
\end{equation}
where $\delta(\cdot)$ is the Dirac Delta function. As proposed in \cite{10}, by thresholding all projection values for individual angles of image using a local threshold and assembling all binarized projections, the Radon barcode for image $I$ can be obtained (Figure \ref{fig:RBC}). A straightforward method to calculate the threshold is taking the median of all non-zero projections values and then binarizing each projection. In order to receive the same length barcodes, all the images are resized into $R_N\times C_N$ images. (i.e., $R_N=C_N=2^N,n\in\mathbb{N}^+$)

\subsection{Gabor Filters}
Gabor filters are bandpass filters which due to their optimal localization in both spatial and frequency domain and orientation selectivity can be used in many applications such as texture feature extraction and texture analysis \cite{20}. The 2D Gabor filter consists of a complex exponential centered at a given frequency and modulated by a Gaussian envelope as follows \cite{5}:
\begin{equation}
G(x,y)=\frac{\omega^2}{\pi \gamma}\exp{\left(-\frac{x'^2+\gamma^2\gamma'^2}{2\sigma^2}\right)\exp{(j2\pi \omega x'+\phi)}},
\end{equation}
where $x'=x\cos(\theta)+y\sin(\theta)$, $y'=-x\sin(\theta)+y\cos(\theta)$, $\omega$ is the central frequency of sinusoid, $\theta$ indicates the orientation of Gabor filter, $\phi$ phase offset in degree, $\sigma$ represents the standard deviation of Gaussian kernel, and $\gamma$ is the spatial aspect ratio. After constructing filter bank by varying the direction and scale, to obtain the Gabor-filtered image $\psi(x,y)$ of a given image $I$, the convolution of each Gabor window in Gabor filter bank with Image  is computed by:
\begin{equation}
\psi_{(u,v)}(x,y)=\sum_s\sum_t I(x-s,y-t)*G_{(u,v)}(s,t),
\end{equation}
where $u=0,1,\dots,U-1$ and $v=0,1,\dots,V-1$. $u$ and $v$  define the scale and orientation of the Gabor filter and $U$ and $V$ are the number of scales and orientation, respectively.  $s$ and $t$ are the size of each Gabor filter bank.

\subsection{Support Vector Machines}
The SVM is a supervised learning algorithm used for pattern classification. Given a set of training samples each belongs to one of two categories in an $n$ dimensional space, the aim of a linear SVM is to construct a separating hyperplane that has the largest distance to the nearest training data point of any class (maximizing the margin). Given training samples $\{x_1,x_2,\dots,x_n\}$ and class label $y_i=\{-1,+1\}$ for sample $x_i$ the general form of decision boundary for linear separable samples to classify all the points correctly is as \cite{21} $y_i (\mathbf{w}\cdot \mathbf{x}_i+b)\geq 1$, $\forall i\in\{1,2,\dots,n\}$, where $\mathbf{w}$ is a weight vector and $b$ is a scalar. The decision boundary can be found by solving the following constrained optimization problem $\min \frac{1}{2} || \mathbf{w}||^2$. 
The non-separable cases can be solved by projecting the input data space to high-dimensional space through a non-linear function $\Phi$. Suppose there exists a kernel function $K(x,y)=\Phi^T(x)\Phi(y)$ for a given test pattern, its label can be estimated by the [22] : 
\begin{equation}
g(x)=\textrm{sgn} \left( b + \sum_i \alpha_i y_i \Phi^T(x_i)\Phi(x)\right).
\end{equation}

Some common kernels include polynomial, $K(x_i,x_j)=(x_i\cdot x_j+1)^d$, and radial basis function, $K(x_i,x_j)=\exp(-\gamma || x_i-x_j||^2)$. However, SVM was originally designed for binary classification but several algorithms such as ``one-against-one'', ``one-against-all'' and DAGSVM have been proposed to extend it for multi-class classification by combining the several binary classifiers \cite{23}. 

\section{Proposed Method}
In this paper, we used the algorithm as proposed in \cite{5a} to extract Gabor-Radon barcode  features (GRBFs) but we also extracted Gabor Radon features (GRFs) by adding new variable to the algorithm (see Algorithm \ref{alg:GaborRadon}). GRFs are extracted by taking the magnitude or absolute value of each $\psi_{(u,v)}(x,y)$ and then converted to a vector. We used the GRFs to train the SVM and Gabor Radon Barcode features GRBFs to retrieve  images.  Algorithm \ref{alg:GaborRadon} shows how the Gabor filter banks are applied to the Radon image (the sinogram), and GRF and GRBF are constructed. In order to receive the same length for GRF and GRBF for all images in the data base, all  images are resized into $R_N\times C_N$.

\begin{algorithm}[t]
\caption{Extracting Gabor Radon and Gabor Radon barcode  features according to [5]. Highlighted code (line 13) shows the extended functionality for later classification. }
\begin{algorithmic}[1]
\label{alg:GaborRadon}
\STATE Initialize GRF$_i$ and GRBF$_i\leftarrow \emptyset$
\FOR{all images $I_i$}
	\STATE $R_N\!=\!C_N\!\leftarrow 128$
	\STATE $I_i \leftarrow$ Normalize($I_i , R_N , C_N$)
	\STATE Set number of projection angles $N_\theta$
	\STATE $I_\textrm{Radon,i}\leftarrow$ RadonTransfom($I$)
	\STATE $I_\textrm{Radon} \leftarrow $ resize($I_\textrm{Radon,i}$, [32 , 32])
	\FOR{$\forall u$\textcolor{black}{$\in \{1,...,N_u\}$} and $\forall v$\textcolor{black}{$\in \{1,...,N_v\}$}}
		\STATE $\psi_{u,v}(x,y)\leftarrow$ Gabor($I_\textrm{Radon,i}$)
		\STATE $\psi_{(ABS-u,v)} (x,y)\leftarrow |(\psi_{(u,v)} (x,y)|$
	\textcolor{black}{\STATE Resample $\psi_{(ABS-u,v)}(x,y)$ with $d_1\!\times\!d_2$ coefficients}
		\STATE GRI$_{u,v,i} \leftarrow$ ReshapeToVector($\psi_{(ABS-u,v)}(x,y)$)
		\STATE \textbf{Gabor-Radon fetaures:\\ GRF$_i\leftarrow$ append(GRF$_i$,GRI$_{u,v,i}$)}: 
		\STATE Get threhsold: $T_{u,v,i}\leftarrow$ FindMedian($GRI_{u,v,i}$)
		\STATE Binarize: $B_{u,v,i}\leftarrow$   Find(GRI$_{u,v,i} \geq T_{u,v,i}$)
		\STATE Append barcode: \\GRBF$_{i}\!\leftarrow\!$ AppendRow (GRBF$_{i},B_{u,v,i}$)
	\ENDFOR
\ENDFOR
\end{algorithmic}
\end{algorithm}

For speeding-up the image retrieval process we split the CBIR system into two stages: classification and retrieval. We combine two ideas: 1) using the same feature for a SVM classification before converting it to a barcode (as proposed in \cite{13}), and 2) combining Gabor and Radon features (as proposed in \cite{5,5a}).  

\textbf{Classification Stage --} In a training stage, we have the following steps:  1) In order to obtain the same-length feature vectors for all  images in the dataset, all the images are resized to small size 128$\times$128, 2) Radon transform is applied on resized images to produce a sinogram, 3) Radon image is then resized into 32$\times$32, 4) Gabor filters are convoluted with the images in step 3 to extract Gabor-Radon features based on Algorithm \ref{alg:GaborRadon}, 5) training the SVM using the extracted features. 

\textbf{Retrieval Stage --} In this stage, we applied the first five steps of the classification stage to all of the images in the training samples and every query image in the test samples. Then GRBFs are extracted and saved. For a given query image, first SVM assigns a class to the images based on its GRF, then according to the classified image label, its GRBF is compared with all GRBFs within that class based on Hamming distance. We then applied $k$-NN with $k=1$ to retrieve the most similar image to the query image.

\section{Experiments and Results} 
\subsection{Image Test Data}
\begin{figure}[!t]
\begin{center}
\includegraphics[width=0.75\columnwidth]{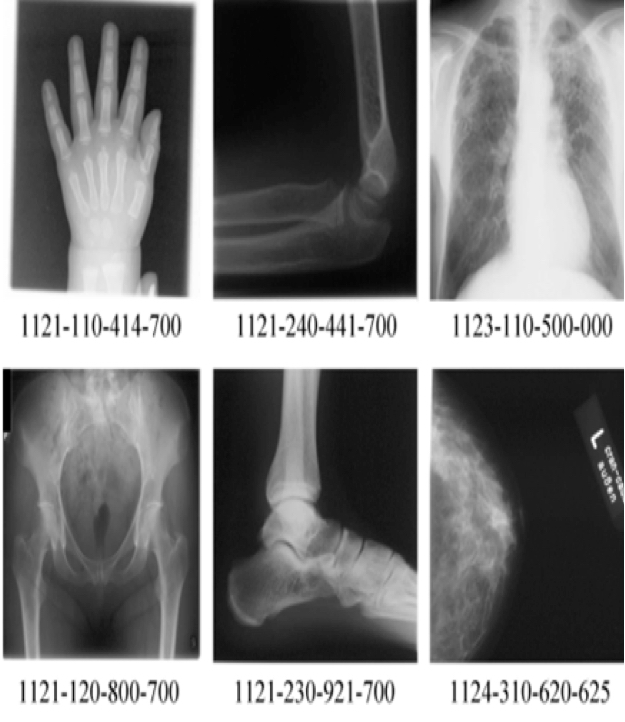}
\caption{Sample images from IRMA dataset  with their IRMA codes.}
\label{fig:irmasamples}
\end{center}
\end{figure}

In this paper the image Retrieval in Medical Applications (IRMA) database \cite{24} the benchmark dataset from \emph{ImageCLEFmed09}, a retrieval challenge in a collection of medical images, is used to validate the proposed method. The IRMA database is a collection of more than 14,000 x-ray images (radiographs) randomly collected from daily routine work which is used for training and testing. All the images were classified in 57 categories and annotated with the IRMA code. The IRMA codes (manually created by several clinicians), which is a string of 13 characters within the set of {0,É,9,a,É,z}, contain information on technical, biological and diagnostic traits of the image in a structured manner: TTTT-DDD-AAA-BBB. Figure \ref{fig:irmasamples} shows some sample images in the IRMA dataset with their corresponding IRMA code.

\begin{figure}[!t]
\begin{center}
\includegraphics[width=0.99\columnwidth]{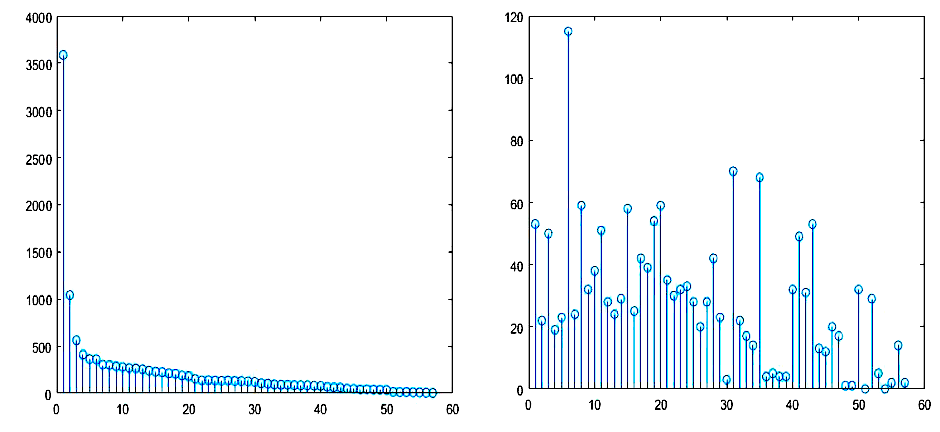}
\caption{Data imbalance in IRMA images: class distribution of training (left), and testing samples (right).}
\label{fig:imb}
\end{center}
\end{figure}

In this dataset, a total of 12,677 x-ray images is used for training and the remaining 1,733 images are considered as testing data. The distribution of data in training and testing images shows that there exists considerable imbalance in IRMA dataset (Figure \ref{fig:imb}).

\subsection{Image Retrieval Error Evaluation Metric}
To evaluate the performance of the proposed algorithm, we used a formula provided by ImageCLEFmed09 to compute the error between the IRMA codes of the testing image and the first hit retrieved by the algorithm, and then summed up the error for all testing images. The error $E$ can be defined as follows \cite{25}:
\begin{equation}
E=\sum_{i=1}^n\frac{1}{b_i}\frac{1}{i}\eta(l_i,\hat{l}_i),
\end{equation}
where $b_i$ is the number of possible labels at position and is the $\eta(\cdot)$ decision function delivering 1 for wrong label and 0 for correct label. We used The Python implementation provided by ImageCLEFmed09 to compute the errors.

\subsection{Image Classification performance}
In the IRMA dataset, there are samples in both training and testing datasets that do not belong to any category. These samples were ignored. Among 12,677 IRMA training images, 12,631 could be used for training and among 1,733 IRMA testing images, 1,639 images were used for testing. We chose the RBF kernel to train the SVM. The parameters of the kernel are set like in \cite{13}.  The ``one- against-one'' approach which is a very robust and accurate method was used to train the multi-calss SVM \cite{23}. In our simulations, the LIBSVM package with Radial basis SVM kernel was used. The classification accuracy of SVM is defined as
\begin{equation}
\textrm{Accuracy}=\frac{|\textrm{correctly predicted labels}|}{|\textrm{testing data}|}.
\end{equation}

\subsection{Results} 
We run the algorithm for different numbers of Gabor filters (12,20,24,48) with a window size of 23$\times$23 and for different numbers of projection angles. Table \ref{tab:details} shows the results. In this table $A$ and $E_\textrm{total}$ refer to the accuracy and total error, respectively, obtained over all images in the testing samples. Also the vector dimension (VD) of the each feature vector has been also listed which can be computed as \cite{5} $\frac{MNN_g}{d_1 d_2}$ where $M$ and $N$ are the size of the Radon transformed image, $N_g$ is the number of Gabor filters and  $d_1\times d_2$ downsampling coefficients (algorithm 2). For example,  for $M=N=32$, $N_g=12$, and $d_1=d_2=4$, we get VD$=768$. 

\begin{table*}[htb]
\centering
\caption{Classification accuracy ($A$), total retrieval error ($E_\textrm{total}$) and vector dimension (VD) for different number of Gabor filters and projections, $n_p$.}
\label{tab:details}
\begin{tabular}{l|ll|ll|ll|l}
&         &    $n_p=8$    &         &  $n_p=16$      &         &  $n_p=32$      &      \\
Gabor Filter Bank                                                               & $A$     & $E_\textrm{total}$ & $A$     & $E_\textrm{total}$ & $A$     & $E_\textrm{total}$ & VD   \\ \hline
GBF(3,4,23,23)                                                                  & 59.31\% & 294.74 & 57.41\% & 283.26 & 57.29\% & 284.19 & 768  \\
GBF(4,3,23,23)                                                                  & 59.55\% & 265.67 & 59.67\% & 268.43 & 59.37\% & 267.65 & 768  \\
GBF(4,5,23,23)                                                                  & 58.21\% & 273.66 & 61.44\% & 251.65 & 61.88\% & 248.03 & 1280 \\
GBF(5,4,23,23)                                                                  & 58.27\% & 277.59 & 60.10\% & 258.18 & 60.28\% & 255.35 & 1280 \\
GBF(4,6,23,23)                                                                  & 57.29\% & 275.85 & 60.22\% & 257.95 & 59.79\% & 260.78 & 1536 \\
GBF(6,4,23,23)                                                                  & 58.45\% & 277.31 & 60.10\% & 258.37 & 60.16\% & 257.09 & 1536 \\
GBF(6,8,23,23)                                                                  & 52.96\% & 310.98 & 54.61\% & 291.15 & 54.42\% & 290.67 & 3072 \\
GBF(8,6,23,23)                                                                  & 56.99\% & 282.83 & 58.39\% & 268.60  & 58.39\%   & 267.72 & 3072 \\ \hline
\end{tabular}
\end{table*}

The best result was achieved for Gabor filter bank with 20 filters and 32 projections angles. The total accuracy and error for this case are 61.88 \% and 248.03, respectively. In Table \ref{tab:comp}, we compared our method with other published results. The state-of-the-art error score on used dataset is 146.5 by Camlica et al. \cite{11}. In their algorithm, multi-scale LBP features are extracted from saliency based folded image data. The decision how to fold image blocks is critical in their methods and takes time for obtaining the saliency map of images in the database. Our method, in contrast is rather simple and fast. 
\begin{table}[!h]
\centering
\caption{Comparing the IRMA errors. Highlighted results all use Gabor and/or Radon barcodes in some way. The proposed method, inspired by \cite{5,5a} and \cite{13}, delivers the best result among this class of algorithms. }
\label{tab:comp}
\begin{tabular}{ll}
Method                   & $E_\textrm{total}$ \\ \hline
Camlica et al. {[}11{]}   & 146.50       \\
TAUbiomed {[}26{]}       & 169.50       \\
Idiap {[}25{]}           & 178.93      \\
FEITIJS {[}25{]}         & 242.46      \\
\CL \textbf{Proposed Method}          & \CL \textbf{248.03}      \\
VPA SabancIUniv {[}25{]} & 261.16      \\
\CL SVM+RBC {[}13{]}         & \CL 294.83      \\
MedGIFT {[}25{]}         & 317.53      \\
\CL GRIBC(5,16,23,23) [5] & \CL 330.36 \\
\CL RBC16 {[}10{]}            & \CL 470.57 \\ \hline    
\end{tabular}
\end{table}

\begin{figure}[!htb]
\begin{center}
\includegraphics[width=\columnwidth]{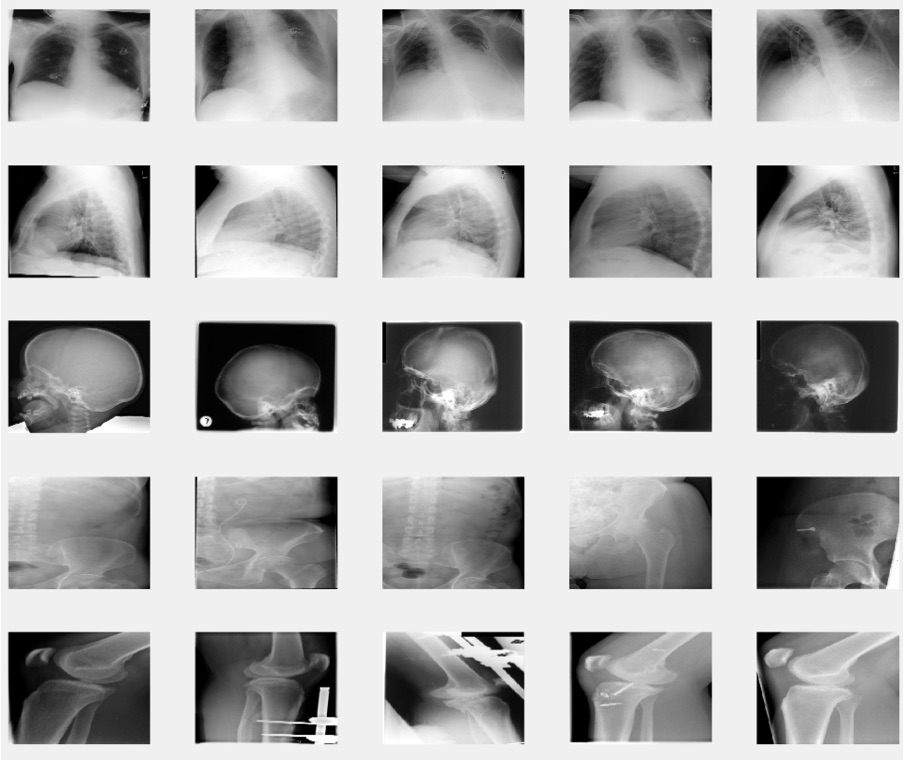}
\caption{Visual results of our method on IRMA dataset. Query image (the first column) and remaining images are most similar images retrived. }
\label{fig:res}
\end{center}
\end{figure}

Figure \ref{fig:res} shows five image retrieval examples. The query image is in the first column and the remaining images are from the training set with the highest similarity with respect to the query image. The results show that our method is able to classify and retrieve similar images reliably.

\section{Conclusion} 
Using Radon and Gabor features for image retrieval have emerged as an interesting research field, specially ``barcodes'', extracted from Radon and/or Gabor features, show some intriguing characteristics for big data. In this paper, we combined two ideas to achieve better results for Gabor-Radon barcodes by using the same features for SVM classification before their binarized form can be used for the final search. Testing with IRMA images, an improvement of state-of-the-art in Gabor-Radon barcodes for image retrieval is apparent.


\begin{thebibliography}{99}
\bibitem{1}    H. Mueller, N. Michoux, D. Bandon, and A. Geissbuhler, ``AÊreviewÊofÊcontent-based image retrieval systemsÊinÊmedical applications clinical benefitsÊandÊfuture directions,'' International journalÊofÊmedical informatics, vol. 73, No. 1, pp. 1-23, 2004.
\bibitem{2}    A. W. M. Smeulders, M. Worring, A. Gupta, and R. Jain, ``Content-based image retrieval at the end of the early years,'' ÊIEEE Transactions on Pattern Analysis and Machine Intelligence,Êvol. 22, No. 12, pp. 1349--1380, 2000. 
\bibitem{3}    Y.Rui, T. S. Huang, and S. F. Chang, ``Image retrieval: Current techniques, promising directions and open issues,'' Journal of Visual Communication and Image Representation, vol. 10 (1), pp. 39-62, 1999. 
\bibitem{4} C.-H. Wei, Y. Li, and C.-T. Li, ``Effective extraction of gabor features for adaptive mammogram retrieval,'' Multimedia and Expo, IEEE International Conference, pp. 1503--1506, 2007.
\bibitem{5} M. Nouredanesh, H.R. Tizhoosh, E. Banijamali, J. Tung. Radon-Gabor Barcodes for Medical Image Retrieval. The 23rd International Conference on Pattern Recognition (ICPR 2016), Cancun, Mexico, December 2016
\bibitem{5a} M. Nouredanesh, H.R. Tizhoosh, E. Banijamali, ``Gabor Barcodes for Medical Image Retrieval,'' IEEE International Conference on Image Processing,2016.
\bibitem{6} T. Tommasi, F. Orabona, and B. Caputo, ``Discriminative cue integration for medical image annotation,'' Pattern Recognition Letters, vol. 29, no. 15, pp. 1996--2002,
2008.
\bibitem{7} T. Tommasi, B. Caputo, P. Welter, M.O. Guld, and T.M. Deserno, ``Overview of the clef 2009 medical image annotation track,'' in Multilingual Information Access Evaluation II. Multimedia Experiments, pp. 85--93, 2010.
\bibitem{8} I. Dimitrovski, D. Kocev, S. Loskovska, and
S. Dzeroski, ``Hierarchical annotation of medical 
images,'' Pattern Recognition, vol. 44, no. 10, pp. 2436--2449, 2011.
\bibitem{9} S. Leutenegger, M. Chli, and R. Y. Siegwart. Brisk: Binary robust invariant scalable keypoints. In IEEE International Conference on Computer Vision, pages 2548--2555, 2011.
\bibitem{10} H.R. Tizhoosh. ``Barcode annotations for medical image retrieval: A preliminary Investigation,'' IEEE International Conference on Image Processing, pp. 818--822, 2015.
\bibitem{11} Z. Camlica, H.R. Tizhoosh, and F. Khalvati, ``Medical image classification via {SVM} using {LBP} features from saliency-based folded data.'' In Machine Learning and Applications (ICMLA), The 14th International Conference on, 2015.
\bibitem{12} M. Verma and B. Raman, ``Center symmetric local binary  co-occurrence pattern for texture, face and bio-medical image retrieval,'' Journal of Visual Communication and Image Representation, vol. 32, pp. 224--236, 2015.
\bibitem{13} S. Zhu andÊH.R.Tizhoosh, ``Radon Features and Barcodes for medical Image Retrieval via SVM,'' International Joint Conference on Neural Networks,2016.
\bibitem{14} A. Sze-To, H.R. Tizhoosh, A.K.C. Wong, ``Binary Codes for Tagging X-Ray Images via Deep De-Noising Autoencoders,'' International Joint Conference on Neural Networks, 2016.
\bibitem{15} J. E. Sklan, A. J. Plassard, D. Fabbri, and B. A. Landman, ``Toward content based image retrieval with deep convolutional neural networks,'' SPIE Medical Imaging. International Society for Optics and Photonics, pp. 94 172C--94 172C, 2015.
\bibitem{16} X. Liu,ÊH.R. Tizhoosh,ÊJ. Kofman, ``Generating Binary Tags for Fast Medical Image Retrieval Based on Convolutional  Nets and Radon Transform'', IEEE International Joint Conference on Neural Networks, 2016.
\bibitem{17} Johann Radon. ``On the determination of functions from their integral values along certain manifolds,'' Medical Imaging, IEEE Transactions, vol. 5, no. 4, pp. 170--176, 1986.
\bibitem{18} P. Cui, J. Li, Q. Pan, H. Zhang, ``Rotation and scaling invariant texture classification based on Radon transform and multiscale analysis,'' Pattern Recognition Letters, Vol.27, No. 5, pp. 408--413, 2006.
\bibitem{19} S. Arivazhagan, L. Ganesan, S. Padam Priyal, ``Texture classification using Gabor wavelets based rotation invariant features,'' Pattern Recognition Letters, Vol. 27, No. 16, pp. 1976--1982, 2006.
\bibitem{20} J. Zhang T. Tan, and L. Ma. ``Invariant texture segmentation via circular Gabor filters,'' Pattern Recognition, Proceedings. 16th International Conference on. Vol. 2, 2002.
\bibitem{21} V. Vapnik and C. Cortes. ``Support-vector networks,'' Journal Machine learning, Vol.20, No. 3, pp. 273--297,1995.
\bibitem{22} ÊK. Crammer, and Y. Singer, `` On the Algorithmic Implementation of Multiclass Kernel-based Vector Machines, '' Journal of Machine Learning Research, Vol.2, pp. 265--292,2011.
\bibitem{23} C.-W. Hsu and C.-J. Lin, ``A comparison of methods for multiclass support vector machines,'' Neural Networks, IEEE Transactions, Vol.13, No. 2, pp. 415--425, 2002.
\bibitem{24} W. R. Hersh, H. Mueller, and J. Kalpathy-Cramer, ``The imageclefmed medical image retrieval task test collection,'' J. Digital Imaging, vol. 22, No. 6, pp. 648--655, 2009.
\bibitem{25} T. Tommasi, B. Caputo, P. Welter, M.O. Guld, and T.M. Deserno, ``Overview of the clef 2009 medical image annotation track,'' In Multilingual Information Access Evaluation. II, vol. 6242,  pp. 85--93, 2010.
\bibitem{26} E. Konen M. Sharon U. Avni, H. Greenspan and J. Goldberger. ``Xray categorization and retrieval on the organ and pathology level, using
patch-based visual words'', Medical Imaging, IEEE Transactions on, volume 30, pages 733--746, 2010.
\end{thebibliography}
\end{document}